\documentclass[10pt, conference, compsocconf]{IEEEtran}
\usepackage[linesnumbered,ruled]{algorithm2e}
\newtheorem{theorem}{Theorem}
\newtheorem{definition}{Definition}
\usepackage{tikz}
\usepackage{graphicx}
\usepackage{pgfplots}
\usepackage{amsmath}
\ifCLASSINFOpdf
\else
\fi
\begin{document}
%
\title{Approximating Permutations with Neural Network Components for Travelling Photographer Problem}


\author{\IEEEauthorblockN{Chong Sue Sin}
\IEEEauthorblockA{Institute for Interdisciplinary Information Sciences (IIIS) \\
Tsinghua University\\
Beijing, China\\
Email: z18516937829@163.com}
}


%


\maketitle

\begin{abstract}
Most of the current inference techniques rely upon Bayesian inference on Probabilistic Graphical Models of observations and do predictions and classification on observations. However, there is very little literature on the mining of relationships between observations and building models of the relationship between sets of observations or the generating context of the observations. Moreover, event understanding of machines with observation inputs needs to understand the relationship between observations. Thus there is a crucial need to build models and develop effective data structures to accumulate and organize relationships between observations. Given a PGM model, this paper attempts to fit a permutation of states to a sequence of observation tokens (The Travelling Photographer Problem). We have devised a machine learning inspired architecture for randomized approximation of state permutation, facilitating parallelization of heuristic search of permutations. As a result, our algorithm can solve The Travelling Photographer Problem with minimal error. Furthermore, we demonstrate that by mimicking machine learning components such as normalization, dropout, and lambda layer with a randomized algorithm, we can devise an architecture that solves TPP, a permutation NP-Hard problem. Other than TPP, we can also provide a 2-Local improvement heuristic for the Travelling Salesman Problem (TSP) with similar ideas.

\end{abstract}


%
\IEEEpeerreviewmaketitle

\section{Introduction}
Given a macro-level phenomenon, we often have models of how likely an observation will occur under a macro-level context. However, when bombarded with a plethora of time-series observation trails, can we simulate phenomenon level simulated rollout to reconcile the macro-level model with observation trails? This effort will provide context to observation trails and ease reasoning about often unorganized time-series observation trails. A macro-level phenomenon often does not directly dictate the occurrence of any single observation, often only probabilistically correlated with each observation and can be well-modelled with Bernoulli distribution. 
$$observation_i | state_m \sim Bernoulli(p_i^{(m)})$$

To solve this problem, we draw inspirations and ideas from successful machine learning models such as activation function, layer normalization and working in embedded space to build a permutation approximation architecture. Our permutation approximation algorithm has proven to be a success on synthetic data. The contribution of this paper can be summarized as follows:

\begin{itemize}
    \item Devised an approximation and randomized architecture which incorporates randomness from Bernoulli simulation in a compact way
    \item Devised problem-driven architecture which can learn from the structure of the problem from the model, giving prediction with 1 data point in 1 shot
    \item Devised an algorithm to solve the Bernoulli stochastic permutation problem.
    \item Potentially for parallelizing heuristic search for permutations
    \item Devised a highly accurate data-specific dropout function
\end{itemize}
We organize the rest of this paper as follows. Section 2 is a literature review, followed by defining the Travelling Photographer Problem in Section 3. An overview of our algorithm for Travelling Photographer is introduced in Section 4, followed by an elaboration on pseudo-state rollout in Section 5. Section 6 elaborates upon the simulated attention mechanism and customized dropout layer for our architecture. Section 7 explores Machine Learning Architecture for Approximation (MLAA) as a template for the real-valued version of the Travelling Photographer Problem. We perform ablation studies on both synthetic data in section 8 and the actual dataset in section 9. In section 10, we customize another MLAA to solve the Travelling Salesman Problem, demonstrating the generalizing capabilities of the architecture across problem types. Finally, we list the method's limitations in section 11, and the paper concludes in section 12. 

\section{Related Work}
This section will introduce related works in two areas, attention mechanism and Travelling Salesman Problem (TSP).
\subsection{Attention Mechanism}
Attention has taken the NLP world by storm with the seminal work "Attention is All You Need" [39]. Many extensions to the attention mechanisms are evident in the number of variants of Transformers, such as Set Transformer[31], Longformer, and InsertionTransformer. Attention can easily be expanded to solve problems other than NLP; it has also demonstrated capabilities in solving routing problems and image captioning by focusing its attention on some internal subsets of the data. The Transformer's self-attention mechanism is a weighted average importance score of its internal words concerning a particular word. Attention performed much better on various machine translation task benchmarks, BLEU than traditional statistical machine translation. Since then, many works have studied the structure and kernel of attention [36], claiming that the effectiveness of attention is due to its entangling of positional embedding and word embedding. However, recent work by Yu, Luo and the team [44] studied the attention mechanism for computer vision tasks. It concluded that replacing the attention mechanism with much simpler computational gadgets such as a pooling layer is possible. This view is supported by Microsoft's DeBERTa's [41] super-human performance on SuperGLUE; DeBERTa disentangles positional and word embedding. 

\subsection{Travelling Salesman Problem Solver}
This section will discuss approximation solvers for the travelling salesman problem. In the 1960s, the Christofides algorithm is an approximation algorithm which solves TSP with a worst-case approximation factor of 1.5. Christofides algorithm was initially thought of as a placeholder, soon to be replaced by a more sophisticated algorithm. However, it remained the best worst-case bound approximation algorithm for TSP until today. From the heuristic search side, the long-standing heuristic search solvers like LKH-solver remain better performing than recent efforts to solve TSP with machine learning models despite many recent efforts. LKH solver is a pair improvement strategy which improves upon a particular TSP solution. Recent efforts to use NLP machine learning tools to solve TSP have attempted to cast TSP encoding and decoding as a translation problem or a general heuristic learning problem. Some works have attempted to train a TSP solver with reinforcement learning methods, whereas \cite{impr} trains a soft-actor critic network to learn an improvement heuristic for TSP solutions instead of a search heuristic. Both \cite{strat}, which combined reinforcement learning strategy selector with LKH, and \cite{pret}, which pre-trains on small-cases of TSP, have successfully solved various hard cases of TSP. In the general machine learning for NP-hard problem schemes, powerful NP-hard solvers like ReduMIS for Maximum Independent Set problem and Z3 have been paralleled by recent work that does general tree search under the guidance of a trained neural network. Chen's work \cite{qifeng} uses a 2-improvement local search strategy on a reduced graph, a combination of search and reinforcement learning techniques.

The Travelling Salesman Problem remains a central problem in the graph learning community that is open to this day. 

\section{Problem Formulation}
Given a set of states that are probabilistically related to observations (model) and a string that collates observations from state traversal (observation trail), we attempt to find the permutation of the state which has most likely generated the observation trail. 

\subsection{The Travelling Photographer Problem}
 Given the following paragraph, an observation trail and some states of agent (Bob), the goal is to identify a sequence of states that most likely generated the observation trail.

Suppose a photographer is travelling between different cities, one city per day. In different cities, he takes pictures of different scenes at will randomly. Sometimes different cities have the same scenes. Now that the photographer is back at home, he has forgotten all about his travelling schedule, and he now wants to figure out his travelling schedule (permutation of cities). The photos have a date it was taken printed below; he wishes to use that to find the order in which he visited the cities. The TPP is about identifying a state permutation to a series of noisy observations. 

We summarize the Travelling Photographer Problem as follows; a city can also be interpreted as a state and given a model of cities. 
City A: $$a \sim Ber(0.1); b \sim Ber(0.2)$$ City B: $$b \sim Ber(0.9); c \sim Ber(0.3) $$ City C: $$b \sim Ber(0.8); d \sim Ber(0.5); 
b \sim Ber(0.7)$$ Find a permutation of $\{A, B, C\}$ which has most likely led to observation sequence $\{1;2;2\&4\}$. In other words, we are finding a sequence $s$ for the following equation. 

\begin{align}
   argmax_{s_i \in \{A, B, C\}} &\quad Pr(a|s_0)Pr(b|s_1)Pr(b,d|s_2)    \\                                  &*Pr(s_1|s_0)Pr(s_2|s_1)
\end{align}

\section{Heuristic Search for Permutation}
The algorithm is a heuristic search over the space of permutations on a user-defined embedded space for observations. In essence, the scoring of each permutation is a Hammersley-Clifford theorem-inspired numerical compression and activation of a state rollout. We attempt to embed the features of each observation episode via a numerical embedding function as input into neural approximation architecture and then perform a ranking on the approximated values. We have a customized activation function similar to the tanh function for input state permutation and observation interpretation. 
\begin{algorithm}[h]
\KwIn{Model $model$, State Transition $T$, Observation $obs$}
\KwOut{Best Fit State Permutation $best$}
$fn = $ {\sc Dropout Function} $(model, T, obs)$ \;
$relevant\_perms =  permutations(states, dropout = fn) $\;
\ForEach{perm in $relevant\_perms$}{
    $score = $ {\sc Sequence Scorer} $(obs, perm, T)$ \;
    \If {$score > max\_score$}{
        $best = perm$ \;
        $score = max\_score$ \;
    }
}
\Return{$best$}
\caption{{\sc Permutation ML-Approx}}
\label{algo:perm_ml_approx}
\end{algorithm}

Our architecture incorporates common machine learning layers such as the attention linear layer, normalization layer, activation function, and lambda layer with multiplication operations. In addition, we designed such that optimization space is a monotonically increasing space. If a state is more likely to have produced a set of observations, we give that state-epoch match a strictly higher score than another state-epoch match is less likely to have led to the same set of observations. The algorithm is such that instead of finding a reward from rolling out a state's Bernoulli observations, we try to draw numerical values from a distribution parameterized by the Bernoulli probabilities to replace a rollout. We score each state's rollout with the product of cliques which is proportional to the likelihood of a state producing observations at hand. Since the number of observations per episode is not constant, we need to normalize the products of each episode concerning the number of observations. 

\begin{algorithm}[h]
\KwIn{Observation Trail $observations$, State Permutation $permutation$, State Transitions $prob$}
\KwOut{Heuristic Score}
initilialize $transitions$ and $memory$ arrays \;
\For{Observation episode t}
{
    $stateScore = $ \sc{Pseudo State Rollout}$(permutation[t], observations[t])$\;
    $memory.append(state Score)$\;
    $transitions.append(prob[permutation[t], state[t+1])$\;
}
$sequence\_score = $ {\sc Attention Block} $(prob, rollout, order)$\;
\Return{$sequence\_score$}
\caption{{\sc Sequence Scorer}}
\label{algo:duplicate}
\end{algorithm}

\subsection{Attention Mechanism Simulation}
 We attempt to simulate the attention mechanism that balances state rollout and the likelihood of state permutation sequence. The input layer takes observation-to-state mapping for each epoch and the probability of length-3 sub-sequences of input state permutation. The algorithm also needs to simulate the self-attention mechanism that the input state permutation performs on itself.


\begin{figure}[ht!]
  \includegraphics[width=\linewidth]{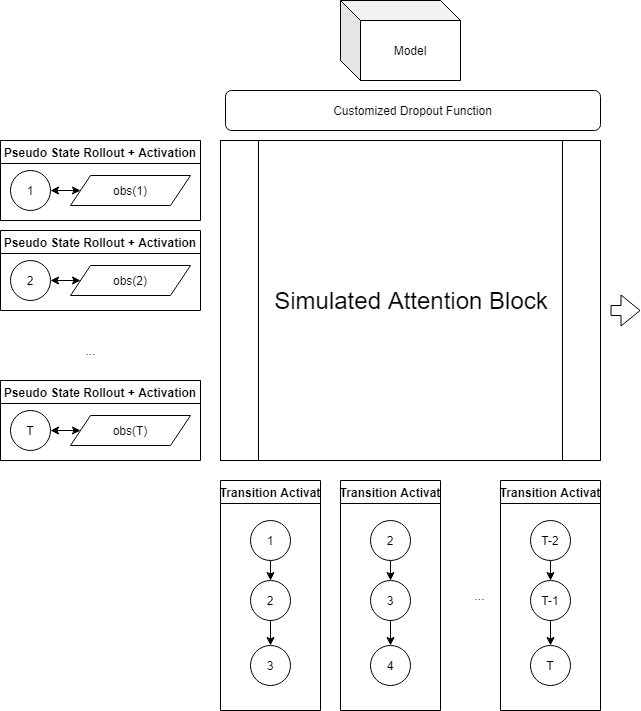}
    \caption{Machine Learning Architecture for Approximation (First Order Attention)}
\end{figure}

\section{Encoding Probabilities from Model}
This paper attempts to score a pseudo-state rollout by repeatedly sampling from a known model, activating the probability that observation occurs with a custom activation function, and returning the product of all the activation values as a heuristic for the likelihood of state generating the set of observation. Hammersley-Clifford Theorem proves the correctness and proportionality.

\begin{theorem} [Hammersley-Clifford Theorem]
Assuming satisfaction of weak positivity condition. Let $U$ denote the set of all random variables under consideration, and let $\Theta, \phi_1, \phi_2, ..., \phi_n \subseteq U$ and  $\psi_1, \psi_2, ..., \psi_m \subseteq U$ denote arbitrary sets of variables. If
$$Pr(U) = f(\Theta)\prod_{i=1}^{n} g_{i}(\phi_i) = \prod_{j=1}^m h_j(\psi_j)$$
for functions $f, g_1, g_2, ..., g_n$ and $h_1, h_2, ..., h_m$, then there exist functions $h_1', h_2',..., h_m'$ and $g_1', g_2',..., g_n'$ such that 
$$Pr(U) = (\prod_{j=1}^m h_j'(\Theta \cap \psi_j))(\prod_{i=1}^ng_i'(\phi_i))$$
In other words, $\prod_{j=1}^m h_j(\psi_j)$ provides a template for further factorization of $f(\Theta)$.
\end{theorem}

\begin{definition} [Weak Positivity Condition]
Functionally compatible densities $\pi_1, ..., \pi_n$ satisfy the weak positivity condition on a set $A \subset S$, if there exists a point $x' \in A$ and a permutation $(r_1, ..., r_n)$ on $\{1, ..., n\}$such that for almost all points $x \in A$ and all $j = 1, ..., n$,
$$
\pi_{r_j}(x_{r_j}' | x_{r_1}, ..., x_{r_{j-1}}, | x_{r_{j+1}}, ..., x_{r_n}) > 0
$$
\end{definition}
First of all, we encode the model probabilities $Pr(o_i|s_j)$'s and $Pr(s_i|s_j)$'s as normal distributions. We then activate drawn samples from the distribution to get edge weights for our intermediate model.
\begin{equation}
    \mathcal{N}(\frac {Pr(o_i|s_j)}{2}, \sqrt{\frac {Pr(o_i|s_j)}{5}})
\end{equation}
In the sections after, we denote activation function by $f$.

\section{Pseudo State Rollout}
{\sc Pseudo State Rollout } is an approximation of the reward function from a state rollout, where the reward function rewards occurrences of observation.
We want to give a score proportional to the likelihood that the given state has produced the observations at hand. Below is our 3-clique scoring scheme.
\begin{equation}
    Pr(o_1, o_2, o_3, o_4|s) \propto \prod_{e_i, e_j, e_k \in \{f(p_1), f(p_2), f(p_3), f(p_4)\}} e_ie_je_k
\end{equation}
The product of all the possible n-cliques of the edge is proportional to the factorized probability that the observations set is associated with a particular state. {\sc Factorized Clique} compressed a state rollout into a score; it performs the function of an embedding layer. We simulate layer normalization with constant multiplications in Pseudo State Rollout and other algorithm subroutines.

\begin{algorithm}[h]
\KwIn{Observation Set at episode t $obs$, State $state$, Model $model$, Clique Size $size$}
\KwOut{Activation Value}
initilialize $value$ as 1 \;
\ForEach{ob in $obs$}
{
    $activations$= \sc{Rollout Activation}$(obs, state, model)$\;
    $combis = combinations(activations, size)$ \;
    $constant = e^{\frac{1}{len(combis)}}$ \;
    \ForEach{clique in combis}
    {
        $value *= multiply(clique) * constant $\;
    }
    
}
$score = \sqrt[3]{value}* len(obs)$ \;
\Return{score}
\caption{{\sc Pseudo State Rollout}}
\end{algorithm}

\begin{figure}
    \centering
    \includegraphics[scale=0.7]{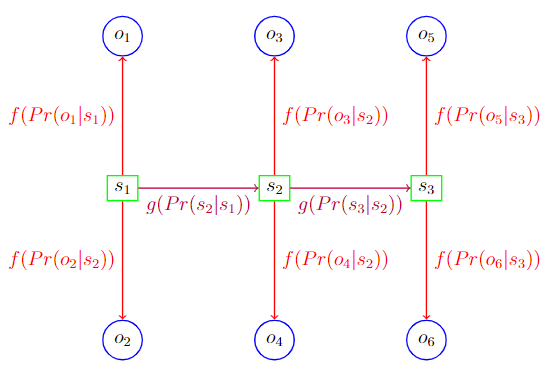}
    \caption{Intermediate Model With Activated Values for Edges}
    \label{fig:my_label}
\end{figure}

\subsection{Activation Functions}
The rollout activation function score is proportional to the likelihood that observation occurs under a particular state. The activation value is thresholded below with a minimum threshold value like a ReLU. The activation function for the simulated self-attention mechanism for state permutations is a minimum instead of a maximum and possesses a threshold from the below feature.

\begin{algorithm}[h]
\KwIn{Observation Set at episode t $obs$, State $state$, Model $model$}
\KwOut{Activation Value}
initilialize $draws$ and $thresh$ values, and $activations$ as empty array \;
\ForEach{ob in $obs$}
{
    \If{ob in $model[state].key()$}{
        $bp = model[state][obs]$ \; 
        \#Bernoulli Probability of Observation Occurring Under State \;
        $distribution = Gaussian(\frac{bp}{2}, \sqrt{\frac{bp}{5}})$ \;
        $samples = sample(distribution, draws)$ \;
        $val = max(thresh, max(samples)$ \;
        $activations.append(val)$ \;
        
    }{
        \Return{empty}
    }
}
\Return{activations}
\caption{{\sc Rollout Activation Function}}
\label{algo:rollout}
\end{algorithm}

Our activation function incorporates randomness in simulation by activating values drawn from the distribution. The usage of activation functions, min, max, and median, is also more intuitive for users; one can direct whether one wants to consider the worst-case, average-case or the best-case in a simulated state rollout. Our architecture has a customized activation function for different input sections of the architecture.

\section{Attention for Transitions and State Rollouts}
We take both activated state permutation and observation graphs as input and simulate embedding them before feeding both data streams into a common multiplicative layer. There are two types of attention mechanisms, self-attention and encoder-decoder attention. 

$$Attention = Softmax(\frac{Q*K}{Scale})*V^T$$
\begin{itemize}
    \item Q = Encoder Layer; K = Decoder Layer; V = Learnt Vector
    \item Self-Attention:            Q = K = V = Source 
    \item Encoder-Decoder Attention: K, V = Source; Q: Target
\end{itemize}

We experimented with different types of encoding and decoding combinations. Proportional to the richness of representation of the mechanism in Natural Language Processing (NLP), the prediction performance also increases accordingly. In {\sc First Order Attention}, we simulated self-attention for input state permutation and performed self-attention on state roll-out keys $\{state[i],obs[i]\}$ as a whole. The decoder is a multiplicative function that attempts to balance the two components.

\begin{algorithm} [h]
\KwIn{Transitions $transitions$, State Rollouts $rollout$}
\KwOut{Heuristic Score}
initilialize $draws$ and $thresh$ values \;
$episodes = length(rollout)$ \;
$transition\_product, rollout\_product = 1, 1$ \;
\For{t in range(episodes - 2)}
{
    $activated = $ \sc{Transition Activation} $(transition[t]*transition[t+1])$ \;
    $transition\_product *= activated * e^{\frac{1}{episodes}}$
}
$transition\_product *= transitions[-1]$ \;
\ForEach{$state$ in $rollout$}
{
    $rollout\_product *= rollout$ \;
}
$score = \sqrt[3]{transition\_product} * rollout\_product$ \;
\Return{score}
\caption{{\sc First Order Attention}}
\label{algo:firstOrder}
\end{algorithm}

In the second-order attention, we process state-observation rollout and likelihood of state sequence in tandem, drawing inspirations from Bidirectional Encoder Transformer (BERT) models. The likelihood of matching the entire sequence is proportional to the score of each of the two time-step matching in the sequence. We performed self-attention on both input sequences and tuple $\{ obs[t], prob(state[t], state[t+1], obs[t+1]\}$ as encoder input to the attention block.

\begin{algorithm} [h]
\KwIn{Transitions $transitions$, State Rollouts $rollout$}
\KwOut{Heuristic Score}'
initialize $transition\_product, rollout\_product = 1, 1$ \;
$episodes = length(rollout)$ \;
\For{t in range(episodes)}
{
    $activated = $ \sc{Transition Activation} $(transition[t]*transition[t+1])$ \;
    $transition\_product =\sqrt[3]{activated}$ \;
    $cur\_rollout = rollout[t]*rollout[t+1]$ \;
    $overall *= transition\_product*cur\_rollout*e^{\frac{1}{episode}}$ \;
}
\Return{overall}
\caption{{\sc 2-Sequence Attention}}
\label{algo:secondOrder}
\end{algorithm}
The above simulation of attention has produced an excellent decoder score. However, the architecture can be confused when there are too many states and a few positions. 
\begin{figure}[h!]
  \includegraphics[width=\linewidth]{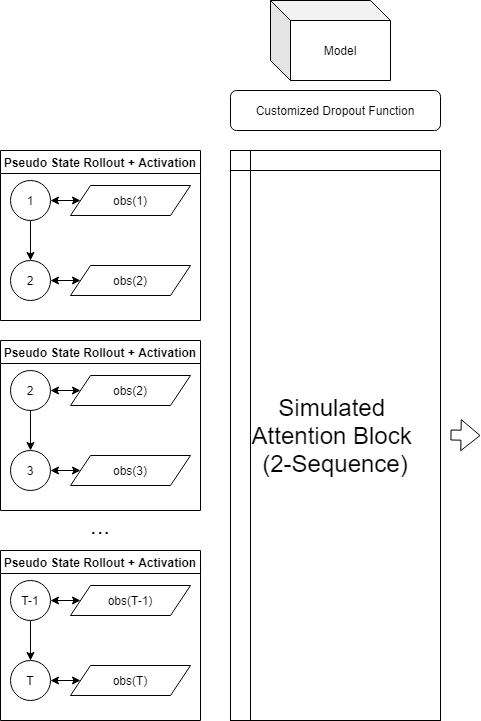}
    \caption{2-Sequence Architecture Diagram}
\end{figure}
This leads us to try to improve the architecture with a denoising filter in the next section.  

\subsection{Simulated Dropout Layer}
Like in knowledge graph problems, we would like to capture how central is a particular state given an observation trail. We distribute the shard of the probability of each of the observations to the states related and propose that transitions are likely to take place between two states with a more significant portion of the probability shard sum. We define centralness as the sum of the probability shard of all transitions connected to it, expressed by the matrix $state_capacity, C$. 

We have customized a dropout function for each input observation trail, and the objective is to include the most central state for each observation sequence. The simulated dropout improves prediction accuracy, effectively speeds up computation, and allows for effective parallelization of the architecture. To achieve this aim, we again borrow ideas from the best practices of machine learning. 

\begin{algorithm}[h]
\KwIn{Model $model$, State Transition $T$, Observations $obs$}
\KwOut{Dropout Function $dropout$}
initilialize $must\_traverse =[]$\;
$state\_capacity, C = $ {\sc Transition Capacity} $(model, obs)$ \;
$M = [C ,CTCT-CT]$ \;
\ForEach{capacity in M}{
$edge\_set = $ {\sc Maximum Spanning Edges} (C) \;
$state\_weights = $ {\sc Weight Sum} $(edge\_set)$\;
$ranking = sort(state\_weights, decreasing = True)$ \;
$must\_traverse.append(ranking.pop())$ \;
}
$dropout = $ lambda s: return $all(must\_traverse)$ in s\;
\Return{$dropout$}
\caption{{\sc Dropout Function}}
\label{algo:dropout_function}
\end{algorithm}

Instead of a random dropout rate, we have an inclusion function which works very well empirically.

\section{Architecture as Template}
Combining features stated in previous sections, here is a compact way of defining the overall algorithm with custom-made activation functions for each section of the algorithm. As a result, users can modify the behaviour of the architecture when performing approximation with varied models and datasets. 

\begin{algorithm}[h]
\KwIn{Model $model$, Rollout Activation Function $rollout$, Transition Activation Function $transition$, Dropout Activation Function $dropout$, Attention Type $attention$}
\KwOut{Architecture}
$dropout\_fn = ${\sc Dropout Function} $(dropout)$ \;
$pseudo\_rollout  = ${\sc Pseudo State Rollout} $(rollout)$ \;
$attention\_block = ${\sc Attention}$(attention, transition)$ \;
$architecture = $ {\sc Permutation ML-Approx} $(dropout\_fn, pseudo\_rollout, attention\_block)$ \; 
\Return{architecture}
\caption{{\sc ML Architecture for Approximation(MLAA)}}
\label{algo:arch_template}
\end{algorithm}

\subsection{Real-Valued Observations}
In this subsection, we consider real-valued input data in the observation trail. The model records the observation under state as a normal distribution or other continuous value distribution functions. The machine learning architecture in {\sc Permutation ML-Approx} serves as a template for solving the problem and can be adapted accordingly for approximating observation trails which include real-valued observations with customized activation functions which work with real-valued observations.

\begin{algorithm}[h]

\KwOut{function}
$function = MLAA(${\sc Real-Valued Rollout Activation}, {\sc Transition Activation Function}, {\sc Real-Valued Dropout Activation Function}, {\sc First Order Attention}) \;
\Return{$function$}
\caption{{\sc Real-Valued Permutation ML-Approx}}
\label{algo:arch_template2}
\end{algorithm}

We input sampled values $L1$ values from the distribution into our architecture to get a proportional score of these $L1$ values, reflecting the likelihood that each state has led to these observations.

\section{Ablation Studies}
We have performed experiments on a synthetic dataset, and the results are satisfying. Furthermore, we have found that this architecture is relatively robust for sequence-to-sequence prediction by performing ablation studies on several hyperparameters of the neural approximation structure. 
\subsection{Experimental Settings}
 Random tours on the state graph generate observation trails on the transition matrix for some episode time steps. The experiment configuration consists of 9 states, and there are at most six observations associated with each state.

The procedure and architecture above has been able to blend well numerically and consistently produce reasonably good predictions. We can see from the plot below that both first and second order compilers improvements improve as the number of episodes increase. 
\begin{tikzpicture}
\begin{axis}[
    title={Ablation on Clique Size},
    xlabel={Episodes of Observation},
    ylabel={$\%$ Error in Predicted String},
    xmin=2, xmax=10,
    ymin=0.0, ymax=0.2,
    ytick={0.0,0.05, 0.1, 0.15, 0.20, 0.2},
    xtick={3, 4, 5,6,7,8,9},
    legend pos=north east,
    ymajorgrids=true,
    grid style=dashed,
]
\addplot[
    color=blue,
    mark=square,
    ]
    coordinates {
    (3, 0.196) (4, 0.092) (5,0.11) (6, 0.078) (7, 0.085) (8, 0.0505) (9, 0.0138) 
    };
    \legend{}{Once (1st Order)}

\addplot[
    color=red,
    mark=triangle,
    ]
    coordinates {
    (3, 0.145) (4, 0.140) (5,0.092) (6, 0.092) (7, 0.075) (8, 0.0454) (9, 0.0023) 
    };
    
\addplot[
    color=green,
    mark=square,
    ]
    coordinates {
    (3, 0.167) (4, 0.087) (5,0.0965) (6, 0.0598) (7, 0.0499) (8, 0.045) (9, 0.0069) 
    };
    \legend{}{1xF-2 (2nd Order)}

\addplot[
    color=orange,
    mark=triangle,
    ]
    coordinates {
    (3, 0.15) (4, 0.125) (5,0.0875) (6, 0.0541) (7, 0.05128) (8, 0.042) (9, 0.0023) 
    };
    \legend{1xF-2 (1st Order), 1xF-3 (1st Order), 1xF-2 (2nd Order), 1xF-3 (2nd Order)}

\end{axis}
\end{tikzpicture}

The algorithm's performance is slightly less satisfactory for cases with fewer episodes because there are more possibilities for each state. In contrast, when the agent visited most states, every state must be bound to a specific position in the output permutation. Therefore will return a sequence for which each state is being matched to a position of the maximum score. We also see that the performance is consistent with the clique size and the number of transitions in each granularity.

\begin{tikzpicture}
\begin{axis}[
    title={Ablation on Repetitions},
    xlabel={Episodes of Observation},
    ylabel={$\%$ Error in Predicted String},
    xmin=2, xmax=10,
    ymin=0.0, ymax=0.16,
    ytick={0.0,0.05, 0.1, 0.15},
    xtick={3, 4, 5,6,7,8,9},
    legend pos=north east,
    ymajorgrids=true,
    grid style=dashed,
]

\addplot[
    color=blue,
    mark=triangle,
    ]
    coordinates {
    (3, 0.145) (4, 0.140) (5,0.092) (6, 0.092) (7, 0.075) (8, 0.0454) (9, 0.0023) 
    };
\addplot[
    color=red,
    mark=square,
    ]
    coordinates {
    (3, 0.071) (4, 0.058) (5,0.05) (6, 0.05) (7, 0.03) (8, 0.016) (9, 0.0) 
    };
    \legend{}{3x (1st Order)}
\addplot[
    color=orange,
    mark=triangle,
    ]
    coordinates {
    (3, 0.15) (4, 0.125) (5,0.0875) (6, 0.0541) (7, 0.05128) (8, 0.042) (9, 0.0023) 
    };

    \legend{1xF (1st Order), 3xF(1st Order), 1xF(2nd Order)}

\end{axis}
\end{tikzpicture}
With repetition and a majority vote on state, the architecture can find unique state permutations for each observation trail, solving the problem with a low error rate. The increase in accuracy from repetition comes from the Law of Large Number. 

\section{Use Case}
Given an actual dataset for which a TPP model is not readily available, we need first to construct a model from the dataset. On Kaggle's Africa Economic Crisis Data, we have defined states with functions on subsets of columns and found state permutations for economic history. Algorithms described in this paper have been able to capture contexts in each case's dataset. For a series of economic observations, {\sc ML-Approx} has been able to interpret economic time-series data with user-defined states using the case's historical dataset as background context. Results from the architecture have provided a flexible and diversified way of interpreting time-series data.

Different conditions have been returned for different cases given the same series of figures. The decoding is generally slightly more optimistic for cases with solid economic history and more pessimistic for cases with weak economic history.

\section{Travelling Salesman Problem Heuristic Scorer}
In this section, we will appropriate MLAA for the Travelling Salesman Problem. We will present a heuristic which attempts to greedily balance between edge cost minimization and every node's potential as a local minimizer. While {\sc TSP Approx} cannot beat a traditional heuristic solver like the LKH solver, we think the scoring heuristic presented below is relatively novel and can potentially speed up a TSP heuristic solver because it helps with filtering. Given two permutations, we score both permutations, the one with the larger score is more likely to be part of an optimal solution. Our heuristic can be helpful if the permutation space is too large and only a subset of states is visible to the algorithm.

\begin{algorithm}[h]
\KwIn{Cost Matrix $cost$, Permutation $perm$}
\KwOut{node\_activation}
initialize $node\_activation, prod, clique\_act = 1, [],1 $\;
\ForEach{i in $perm$}{
    $edge\_cost = \sum_{edge (i, j) \in perm} wij$ \;
    $denom = \sum_{j = \Omega} wij - edge\_cost$ \;
    $prod.append(\frac{edge\_cost}{denom})$
}
$combis = permutations(prod, 3)$ \;
$k = len(combis)^2$ \;
\ForEach{combi in $combis$}{
    $p1, p2, p3 = combi $ \;
    $clique\_act *= p1*p2*p3*e^{\frac{1}{k}} $\;
}
\Return{$\sqrt[3]{clique\_act}$}
\caption{{\sc Node Activation}}
\label{algo:arch_template3}
\end{algorithm}

We simulate the attention mechanism with multiplication operations between node activation and edge activation outputs. Node activation values indicate how much of the local minimizer potential of the node in permutation has been exploited. We select all cliques (triangles) and get a product of those to get a score proportional to the degree to which each node's minimizing potential in the permutation is exploited.

\begin{algorithm}[h]
\KwIn{Cost Matrix $cost$, States $states$, Length $length$}
\KwOut{local minima permutation, score}
$perms = permutations(states, length)$ \;
\ForEach{perm in $perms$}{
    $edge\_activation = ${\sc Edge Activation} $(cost, perm)$ \;
    $node\_activation = ${\sc Node Activation} $(cost, perm)$ \;
    $score = edge\_activation * node\_activation$ \;
    \If{$score \geq best\_score$}{
    $best = perm$ \;
    $best\_score = score$ \;
    }
}
\Return{$best$, $best\_score$}
\caption{{\sc Local Minima}}
\label{algo:arch_template4}
\end{algorithm}

Local Minima is simply a wrapper for simulated attention mechanisms between edge and node activations. For a set of states, it picks a permutation of states which is the local minima permutation of a particular length and returns that. We approximate the global optima by concatenating 3-4 local optima in smaller node subsets, which can drastically speed up computation. The randomness comes from the random distribution of states into a smaller set of nodes. By repeating the experiments, we evaluate new partitions of states and determine whether the best solution it produces is a good fit for the TSP solution. Of course, the greater the score, the more recommendable is the current portion of the solution.

\begin{tikzpicture} 
\begin{axis}[
    title={Ablation of TSP Approx},
    xlabel={lg (Number of Repetitions)},
    ylabel={Approximation Factor},
    xmin=1, xmax=5,
    ymin=1., ymax=1.3,
    ytick={},
    xtick={1,2,3,4,5},
    legend pos=north east,
    ymajorgrids=true,
    grid style=dashed,
]

\addplot[
    color=blue,
    mark=triangle,
    ]
    coordinates {
    (1, 1.26) (2, 1.176)  (3,1.07)  (4,1.01) (5, 1.0)
    };
    \legend{9 cities x 3}

\end{axis}
\end{tikzpicture}

We appropriated MLAA to solve TSP, and we obtained performance as follows. We can see that our heuristic can figure out an optimal solution very early on. 

\subsection{Simulated Attention as Improvement Heuristic}

Our simulated Attention for TSP can also be appropriated as an improvement heuristic to give satisfactory results for TSP20, and TSP100 problems, often with an error rate of less than 10\%, returning the optimal solution at times. TSP solutions from our methods can compare to modern heuristic solvers for TSP for TSP20 and TSP100. The LKH heuristic is a 3-local heuristic, whereas ours is a 2-local heuristic. Our heuristic might possess a massive potential for parallelization on a large scale. We leave the empirical evaluation of this method on larger cases of TSPs as future work. 

\section{Limitations and Future Work}
The method discussed in this paper requires careful balancing of weighing factors with normalization constants; the outcome is sensitive to changes in normalization constants. Future works should address methods to robustify the simulated attention approach to develop an architecture that allows for both transparencies and does not rely heavily on normalization for correct results. It will be a massive improvement if the weights of different components in the input can be explored in an unsupervised manner with randomized optimizers like Stochastic Gradient Descent instead of relying on hard-coded architecture parameters. Future work can also choose to simulate skip connections and prompting. 

On top of that, we would also like to expand our algorithms for more extensive cases of Travelling Photographer Problems and Travelling Salesman Problems with practical parallel implementations. 

\section{Conclusion}
In conclusion,  MLORA can provide the explainability feature that neural networks are still unable to provide. Thus, we proposed an architecture for randomized algorithms inspired by machine learning to perform approximations to provide more human interpretable structures. Furthermore, we can effectively utilize a simulated attention mechanism to solve other permutation problems. By combining attention to both edge and nodes independently yet simultaneously, we can further push the boundaries of randomized algorithms to solve probabilistic permutation (NP-Hard problems) with heuristics that take advantage of both repetition and randomness. Such implementations help us understand the numerical landscape of the attention mechanism better; it is also a heuristic for paralleling and randomizing the search for permutations. 

For both Travelling Photographer Problem and Travelling Salesman Problem, our architecture can produce satisfactory results for small test cases, which serves as a proof of concept for the feasibility of our machine learning inspired architecture for approximation to solve NP-Hard problems.

\appendix
In the appendix, we list other components of the Machine Learning Architecture. {\sc Transition Capacity} reflects the centralness of each state in an observation sequence. 
\begin{algorithm}[h]
\KwIn{Model $model$, Observation $obs$}
\KwOut{Transition Capacity Matrix $cap$}
initilialize $cap =[]$\;
\ForEach{ob in $obs$}{
    \ForEach{pair of states which can produce ob $, (s_a, s_b)$}{
        $activated =$ {\sc Dropout Activation} $(s_a, s_b, ob)$ \;
        $cap[s_a][s_b] += activated$ \;
        $cap[s_b][s_a] += activated$ \;
    }
}
\Return{$Upper\_Triangle(cap)$}
\caption{{\sc Transition Capacity}}
\label{algo:transition_capacity}
\end{algorithm}
We construct an adjacency graph of states, which edge weights is proportional to likelihood of a transition taking place between the nodes given the observation sequence. 

\subsection{Activation Functions}
In this subsection, we list the activation functions for transition and dropout function. We want to capture the best case for two-step transition, this is reflected by the max function in {\sc Transition Activation Function}. 
\begin{algorithm}[h]
\KwIn{Transition Probability $prob$}
\KwOut{Activation Value}
initilialize $draws$ and $thresh$ values \;
$distribution = Gaussian(\frac{bp}{2}, \sqrt{\frac{bp}{5}})$ \;
$samples = sample(distribution, draws)$ \;
$activation = max(thresh, min(samples))$ \;
\Return{activation}
\caption{{\sc Transition Activation Function}}
\label{algo:transition}
\end{algorithm}

For dropout activation function, we want to capture the likelihood that on average how likely is the node going to be involved. This is reflected by the median function choice. 
\begin{algorithm}[h]
\KwIn{Model $model$, State A $state_a$, State B, $state_b$, Observation $obs$}
\KwOut{Activation Value}
initilialize $draws$ and $thresh$ values $=5, 0$\;
$p_a = model[state_a][obs]$ \;
$p_b = model[state_b][obs]$ \;
$distribution_a = Gaussian(\frac{p_a}{2}, \sqrt{\frac{p_a}{5}})$ \;
$distribution_b = Gaussian(\frac{p_b}{2}, \sqrt{\frac{p_b}{5}})$ \;
$samples_a = sample(distribution_a, draws)$ \;
$samples_b = sample(distribution_b, draws)$ \;
$value = median(samples_a)* median(samples_b)$ \;
$activation = max(thresh, value)$ \;
\Return{activation}
\caption{{\sc Dropout Activation Function}}
\label{algo:dropout}
\end{algorithm}
We think these choices of functions reflects our intention to capture the best, average or worse case for probabilistic events. It might help us understand why activation functions are crucial for neural networks.

\subsection{Real-Valued TPP}
For real-valued case, we again have customized activation functions, which help us find an intermediate mapping between raw numerical values and distance between average cases of draws to get a better overall heuristic score. 

\begin{algorithm}[h]
\KwIn{Model $model$, State A $s_a$, State B, $s_b$, Observation $obs$}
\KwOut{Activation Value}
$dist = {energy\_distance(model[s_a][obs], model[s_b][obs])}$ \;
$activation = \frac{1}{dist}$ \;
\Return{activation}
\caption{{\sc Real-Valued Dropout Activation Function}}
\label{algo:rdropout}
\end{algorithm}

For roll-out activation function, we have separate functions for Gaussian distribution case and non-Gaussian distribution case.

\begin{algorithm} [h]
\KwIn{Model $model$, Observation at episode t $obs$, State $state$}
\KwOut{Heuristic Score}
initilialize $thresh, score$ values \;
$indices\_combi = combinations(len(obs), 3)$ \;
\ForEach{$obs\_indices$ in $indice\_combi$}{
\ForEach{ind in $obs\_indices$}
{
$val = obs[ind]$ \;
Observation@State Model, $m = model[state][ind]$ \;
$activated =$ {\sc Model Heuristic} (m, value thresh, gauss) \;
$score *= activated$
}
}
\Return{score}
\caption{{\sc Real-Valued Rollout Activation}}
\label{algo:RVRolloutAct}
\end{algorithm}

\begin{algorithm} [h]
\KwIn{Model $model$, Observation $val$, Minimum Threshold $thresh$, Is Model Gaussian $gaussian$}
\KwOut{Observation Activation}
\uIf{$gausian$ is True}{
$activation = min(z\_val(obs, m), thresh)$ \;
}\uElse{
$l1\_dist = |max(m) - pdf(obs, m)|$ \;
$activation = \frac{1}{l1\_dist}$ \;
}
\Return{$activation$}
\caption{{\sc Model Heuristic}}
\label{algo:ModelHeuristic}
\end{algorithm}

\nocite{*}
\bibliography{References}
\bibliographystyle{plain}

\end{document}